# A lightweight model FDM-YOLO for small target improvement based on YOLOv8

Xuerui Zhang
College of Mathematics and Statistic
Chongqing Univerity, Chongqing

*Small targets are particularly difficult to detect due to their low pixel count, complex backgrounds, and varying shooting angles, which make it hard for models to extract effective features. While some large-scale models offer high accuracy, their long inference times make them unsuitable for real-time deployment on edge devices. On the other hand, models designed for low computational power often suffer from poor detection accuracy. This paper focuses on small target detection and explores methods for object detection under low computational constraints. Building on the YOLOv8 model, we propose a new network architecture called FDM-YOLO. Our research includes the following key contributions: We introduce FDM-YOLO by analyzing the output of the YOLOv8 detection head. We add a high-resolution layer and remove the large target detection layer to better handle small targets. Based on PConv, we propose a lightweight network structure called Fast-C2f, which is integrated into the PAN module of the model. To mitigate the accuracy loss caused by model lightweighting, we employ dynamic upsampling (Dysample) and a lightweight EMA attention mechanism.The FDM-YOLO model was validated on the Visdrone dataset, achieving a 38% reduction in parameter count and improving the Map0.5 score from 38.4% to 42.5%, all while maintaining nearly the same inference speed. This demonstrates the effectiveness of our approach in balancing accuracy and efficiency for edge device deployment.*

## I. INTRODUCTION

In the realm of computer vision, object detection is pivotal across numerous sectors, such as autonomous vehicles [1] ,surveillance of traffic scenes [2], boosting intelligent driving systems[3], and aiding search and rescue operations [4]. Precise identification of small objects like pedestrians, cars, motorcycles, bicycles, traffic signs, and signals is fundamental for secure navigation and decision-making in autonomous vehicles and intelligent driving systems [5], Additionally, recognizing small objects contributes to better traffic flow management, pedestrian protection, and comprehensive analysis of traffic scenarios. This skill is vital for the enhancement of urban planning and transport networks [4].

Detecting small objects in images presents significant challenges due to their limited spatial coverage, lower resolution, and less distinctive visual features compared to larger objects. In network architectures like YOLOv8 [6], shallow layers may inadvertently discard critical spatial information necessary for identifying these small objects, leading to data loss. Furthermore, during the feature extraction process, smaller objects can be dominated by larger ones, potentially resulting in the loss of important details essential for precise detection. Addressing these issues is vital for enhancing the accuracy and reliability of object detection in practical applications.

With the reduction in production costs and advancements in flight control technology for drones, these compact and agile devices are increasingly being utilized for intelligent traffic monitoring . Drones typically operate at higher altitudes to capture a broader field of view, but this increased distance reduces the apparent size of ground objects. This distance complicates object detection in the captured images. Despite significant progress in object detection, detecting small objects such as pedestrians, motorcycles, bicycles, and vehicles in urban traffic remains challenging due to their varying sizes, diverse shapes, and cluttered backgrounds. This challenge becomes even more pronounced when using limited hardware resources in computer vision and object detection tasks. Balancing detection performance and model size is a critical challenge that needs to be addressed.

To address the challenges of small object detection in drone aerial imagery and traffic scenarios, and to optimize deployment performance, we have developed a novel model based on YOLOv8. We extended the length of the PAN (Path Aggregation Network) and added an additional detection head while removing the large detection head, aiming to more effectively utilize high-resolution spatial details while maintaining a balance in performance. Furthermore, we integrated the Pconv [8] method into the C2f module to further reduce performance overhead during deployment. Additionally, we introduced the EMA[9] (Efficient Multi-scale Attention) mechanism and a dynamic upsampling method into the PAN[10].

The organization of this paper is structured as follows: Section 2 reviews relevant literature and related work. Section 3 elaborates on the proposed improvements to YOLOv8. Section 4 describes the experimental methodology and analyzes the results.

## II. RELATED WORK

The R-CNN, which emerged in 2014, is a two-stage object detection algorithm based on deep learning. This algorithm models object detection as two steps: first, the generation of candidate regions, and second, the classification and regression operations for each candidate region. It uses SVM classification to determine whether a region contains a specific object and regression to obtain the specific location of the object's bounding box. Fast-RCNN optimizes the algorithm's time significantly by allowing each candidate region to share the same neural network. Faster-RCNN [11] further optimizes the candidate region generation algorithm by proposing the RPN network architecture. RPN eliminates the need for the algorithm to rely on traditional feature-based selective search for candidate regions. In 2018, Cascade R-CNN [12] adopted a cascaded detection mechanism, gradually increasing the IOU threshold, enabling the model to provide more accurate predictions at each step. The FPN network introduced a feature pyramid, processing features at different levels differently, thereby improving the model's ability to detect multi-scale targets.

Since the two-stage algorithm models object detection as a method of first generating candidate regions and then classifying and regressing, each step requires independent computational resources, leading to longer times for the two-stage algorithm. This approach often achieved higher accuracy in the early stages but had slower detection speeds. In contrast, the YOLO algorithm is a single-stage object detection algorithm. YOLO directly models the entire detection task as a regression problem, predicting the image's bounding boxes and class probabilities directly after partitioning each image, thus meeting real-time performance requirements.

The YOLO algorithm divides the image directly into grids and predicts a certain number of bounding boxes and confidence scores for each grid. Early versions of YOLO had lower accuracy and were less sensitive to small objects. YOLOv2 used the Anchor mechanism to enhance the detection capability for small objects [13] and supported advanced techniques such as multi-scale training and Batch Norm.

YOLOv3 further improved the detection of small objects by using multi-scale prediction heads. In April 2020, YOLOv4 [14] adopted an enhanced architecture with bag-of-specials integration and used advanced training methods like bag-of-freebies. To enhance model robustness, it performed adversarial attacks on input images and used genetic algorithms for hyperparameter optimization. This model achieved a mean average precision of 43.5% and an AP50 of 65.7% on the COCO dataset [15]. YOLOv5 used updated training strategies, including Mosaic and Cutmix. In June 2022, the Meituan technology team launched YOLOv6, which used EfficientRep for a more efficient network structure. In terms of deployment, YOLOv6 [16] also used parameter renormalization techniques to accelerate model usage during the deployment phase.

As the YOLO series continues to improve, the accuracy of the YOLO algorithm has also increased. YOLOv7 [17] adopted the Extended Efficient Layer Aggregation Network (E-ELAN). By controlling the shortest and longest gradient paths, it allows deeper models to learn and converge more effectively. YOLOv7 proposed a new scaling strategy based on a tandem model, where the depth and width of the blocks are scaled by the same factor to maintain the model's optimal structure.

The backbone network of YOLOv8 adopted the C2F architecture. Today, the YOLO series continues to update [18]. On the general object detection COCO dataset, the larger versions of the YOLO series have achieved high accuracy.

Another hot research direction in object detection is DETR, which introduces Transformer into the object detection task. This method does not require post-processing of non-maximum suppression or the introduction of prior knowledge Anchors. It predicts a fixed number of bounding boxes and defines the labels as a fixed number, using the Hungarian algorithm for bipartite graph matching between the prediction set and the label set to complete detection. This method performs well on large objects but poorly on small objects.

Zhu X proposed a deformable attention mechanism and an iterative bounding box correction method to optimize detection results [19]. Wang Y [20] proposed an anchor-based query method based on prior knowledge, improving model performance. However, the training of DETR models is not easy to converge, especially for small object datasets where the amount of data is often insufficient.

Efficient DETR [21] analyzed various model initialization methods and combined the characteristics of set prediction and dense detection to speed up model training. Li [22] reduced the instability of the bipartite matching mechanism in DETR by using noisy object queries as additional decoder inputs, proposing DN-DETR. DINO [23] proposed a hybrid object query selection method for anchor initialization and a two-forward propagation mechanism for box prediction, providing a contrastive denoising module and adding an additional DN loss, which further improved the detection ability and real-time performance for small objects.

Although the accuracy of the DETR series models is high, their real-time performance is poor. For easier deployment, Lite DETR [24] reduced the complexity of the model through a key-aware deformable attention mechanism, but the computational load itself was not reduced. RT-DETR [25] utilized Vit [26] to efficiently process multi-scale features, providing real-time performance and maintaining high accuracy by decoupling intra-scale interaction and cross-scale fusion.

In addition to the basic model architecture, there are many strategies for small object detection. In terms of loss functions, literature [27] proposed Feedback-driven loss by increasing the weight of small objects in localization loss. Literature [28] believed that IOU-based loss is unfair for small object matching and modeled the bounding box as a Gaussian distribution, using Wasserstein distance to provide scale invariance and smoother position difference processing for small objects.

From the perspective of improving the size of small objects, some studies first magnify the image before detection, using super-resolution methods to improve detection effects. Cui Z [29] integrated a super-resolution self-supervised framework, proposing AERIS for an end-to-end fusion method. Multimodal methods have also been used in object detection research. Literature [30] studied multimodal object detection with RGB and thermal cameras, proposing a probabilistic fusion strategy for different modal information under Bayesian rules and independence assumptions. The literature proposed a sliding window-based object detection method [31], which divides the image into blocks for sequential detection and then merges them, with the merged results far superior to independent detection.

Despite the progress made in existing research, small object detection methods still face challenges in drone aerial photography and traffic scenarios. It is difficult to balance the accuracy, real-time performance, and parameter quantity of a large number of models. Inspired by partial convolution, we have constructed a lightweight Fast-c2f structure. Additionally, the EMA attention mechanism is introduced to reassign feature weights to enhance feature extraction. Unlike other attention mechanisms, it overcomes the limitations of neglecting the interaction between spatial details and the limited receptive field of 1x1 convolution kernels, which restrict local cross-channel interaction and context information modeling. Moreover, the accuracy of the model is improved by using a dynamic upsampling method.

## III. METHOD

In this section, we'll explore how to enhance the network architecture based on YOLOv8. This upgraded network significantly boosts the capability to detect small targets while reducing the model's parameter count by a substantial 40%. Importantly, it achieves this without noticeably increasing inference latency or additional overhead, making it ideal for detecting small targets in low-computing environments. This model is named FDM-YOLO.

Figure 0 illustrates our network architecture. The main improvements include: 1) optimizing the design of the detection head, 2) introducing a lightweight Fast-C2f structure in the PAN section of the model, 3) utilizing dynamic upsampling, and 4) incorporating an EMA attention mechanism for feature fusion.

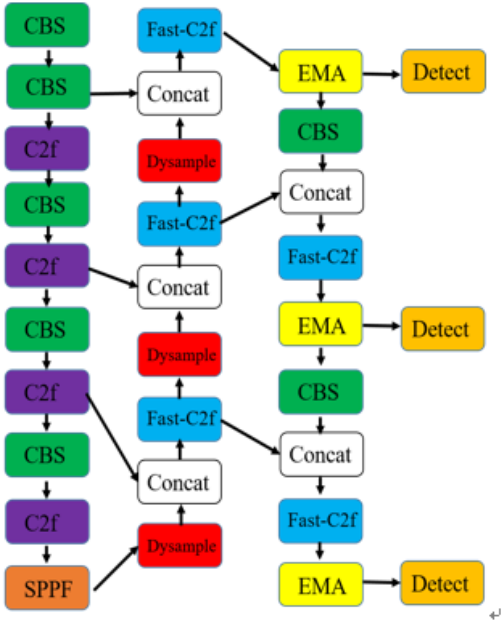

Figure 1 The network architecture of FDM-YOLO

### A. Improve Detect head

In the original YOLOv8, the smallest object detection layer outputs dimensions of (256, 80, 80), which is an 8x downsampling of the original image. This level of downsampling is quite significant for detecting small objects. To enhance our model's ability to detect smaller targets, we extended the length of PAN and FPN and added a new detection layer with a 4x downsampling rate. To maintain performance balance, we removed the largest detection layer.

### B. Fast-C2f

Lightweight deployment has always been an effect pursued by researchers. Since the inception of ResNet, models have mitigated the vanishing gradient problem through residual connections, thereby enabling the construction of increasingly deeper network architectures [30].

However, the deepening of network structures is detrimental to model deployment. The use of lightweight convolutions has also garnered increasing attention from researchers. In this paper, we introduce the lightweight convolution PConv [46] into YOLOv8 to achieve lightweight and reduce the number of model parameters.

The diagram illustrates the computational approach of Pconv, which segments the input feature map by a given scaling factor. One portion of the convolution operates using the standard convolution method, while the other portion is directly replicated. The concatenation of these two parts serves as the final output. This method significantly reduces both the computational load and the memory access volume. In terms of computational load, assuming the dimensions of the output feature map are C*H*W, the computational load of a normal convolution is as follows:

$$\text{FLOPs(Conv)} = c \times h \times w \times k \times k \times c$$

In the above equation, k represents the size of the convolution kernel. In Pconv, the computational load depends on the convolution factor $c_p$, and overall, its algorithmic complexity is:

$$\text{FLOPs(Pconv)} = c_p \times h \times w \times k \times k \times c_p$$

$c_p$ is often defined as $0.25\ c$. Therefore, the overall computational load is reduced to one-sixteenth. In the above calculations, the addition operations have been omitted.

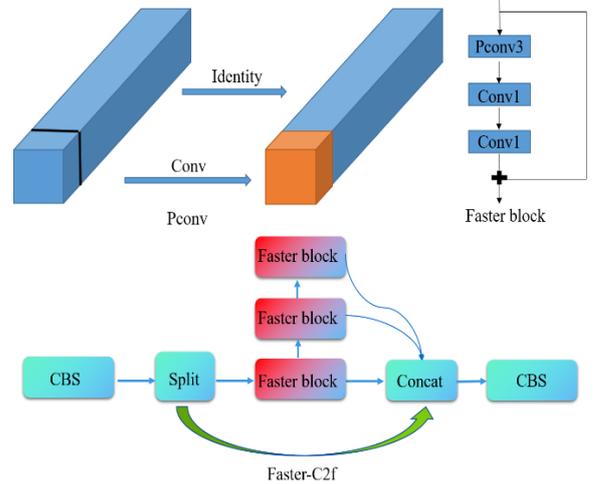

Figure 2 The structure of Pconv and Fast-C2f

By leveraging the combination of Pconv and 1x1 convolutions in series, we can construct a Fast-block, which we then apply within the C2f architecture to create the Fast-C2f module.

In neural networks, the backbone is primarily responsible for extracting core features. To prevent the loss of significant features, the Fast-C2f is not employed in the backbone network. However, during the feature fusion phase of the model, it is utilized to achieve a more lightweight model.

### C. Dynamic upsample

In the YOLO series, the nearest neighbor interpolation method is uniformly employed for upsampling. Another method, bilinear interpolation, estimates the new pixel value by using a weighted average of the color values of the surrounding four pixels, with the weights being the relative distances between the new pixel and the surrounding four pixels. Both of these are classic upsampling methods, and in addition, common methods such as cubic interpolation are also used for upsampling.

However, traditional methods are static upsampling methods, meaning that a rule is given and applied uniformly to all datasets. In reality, there may be inherent differences between different datasets. The upsampling method should possess a dynamic nature.

This paper leverages the Dysample method proposed in the literature [48] to optimize the YOLOv8 object detection model.

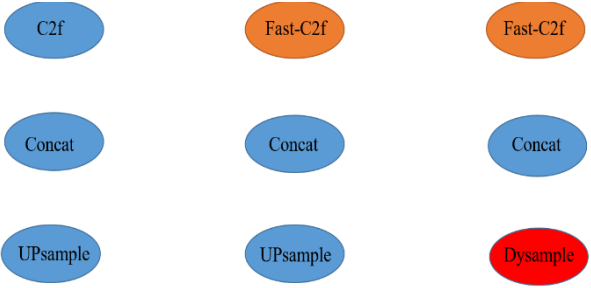

Figure 3 The structure of detection head

*D. Add EMA Attention*

The attention mechanism is a prevalent technique in neural networks, inspired by human visual attention. When processing information, humans do not focus on every detail globally but rather selectively concentrate on certain parts automatically. For an input feature map, it is essential to allow the model to dynamically allocate different weights to various parts, thereby enabling the model to enhance the recognition of significant features.

The EMA method possesses the dual capabilities of cross-channel interaction and cross-spatial learning, and it is more efficient in terms of both performance and effectiveness compared to traditional attention methods. Below is an introduction to the computational process of EMA attention.

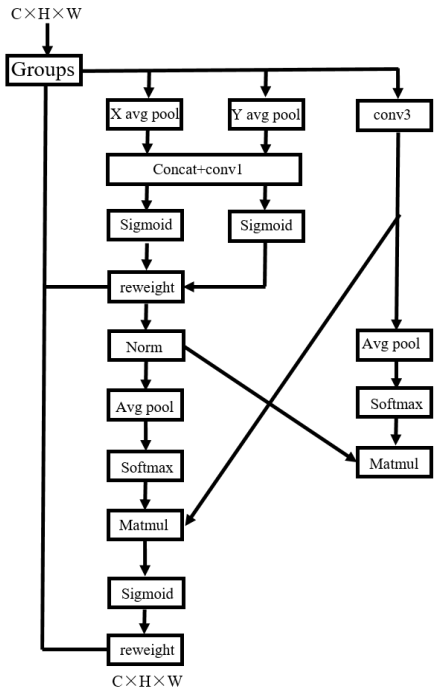

Figure 4 The computational process of EMA attention.

Assuming $T$ is the input feature map, EMA performs average grouping along the channel dimension. We named $T = [T_1, ..., T_n]$, Subsequently, parallel computational operations are conducted for each segment. Each segment simultaneously processes three operations: the first is average pooling along the x-axis, the second is average pooling along the y-axis, and the third is the conventional convolution operation.

The first two parts are concatenated via tensors and then subjected to a 1x1 convolution before being split. After passing through the Sigmoid function, the spatial attention scores are obtained. The original input is then multiplied by these scores to complete the first part of the computation.

The aforementioned process completes the computation of spatial attention. Subsequently, the reweighted Ti and the computed Gi are sequentially subjected to pooling and softmax along the channel dimension to obtain the attention weights in the channel dimension. The attention scores of Ti are multiplied by Gi, and the attention scores of Gi are multiplied by Ti. The sum of these two is then passed through a sigmoid function to become the final attention scores. The product of these attention scores and the original result, after a change in the channel dimension, constitutes the computational result of EMA.

The original version of YOLOv8 did not opt to utilize an attention mechanism, which we have separately integrated into the detection head layer. This incorporation is achieved at a relatively low cost to enhance the detection accuracy. The method of addition is illustrated in the accompanying figure.

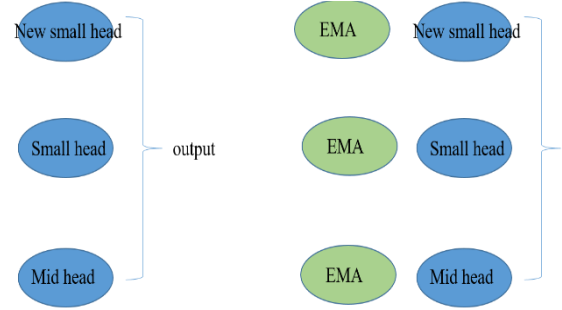

Figure 5 Impore the detection head

## IV. EXPERIMENT

This section begins with an overview of the metrics used to evaluate the performance of models for real-time object detection. Subsequently, the datasets employed for testing are introduced, followed by a detailed explanation of the experimental setup and training strategies. The study uses YOLOv8 as a benchmark to sequentially validate the impact of each innovation on the model. Additionally, the model is compared with other common state-of-the-art (SOTA) methods. Furthermore, this section includes an assessment of the model's performance and relevant discussions.

*A. Evaluation metrics*

To evaluate the detection performance of our improved model, we employ several key metrics: precision, recall, mAP0.5, mAP0.5:0.95, and the number of model parameters. The detailed formulas for these metrics are outlined in this section.

**Precision**: a metric that quantifies the proportion of correctly predicted positive instances (TP) out of all the instances that were predicted to be positive (the sum of TP and FP). Precision is given by the following formula:

$$\text{Precision} = \frac{TP}{TP + FP}$$

**Recall**: This metric calculates the proportion of correctly identified positive samples relative to the total number of actual positive samples, as specified in Equation:

$$\text{Recall} = \frac{TP}{TP + FN}$$

**Average Precision (AP):** represents the area under the precision-recall curve, calculated using Equation 6:

$$AP = \int \text{Precision}(\text{Recall}) d(\text{Recall})$$

**Mean Average Precision (mAP):** represents the mean average precision (AP) value across all categories, reflecting the model's comprehensive detection performance over the entire dataset. This computation is detailed in Equation 7.

$$mAP = \frac{1}{N} \sum_{i}^{N} AP_i$$

where $AP_i$ represents the average precision value for the category index by $i$, and $N$ means the total number of categories in the dataset.

**mAP$_{0.5}$:** the average precision computed with an Intersection over Union (IoU) threshold set to 0.5.

**mAP$_{0.5:0.95}$:** refers to calculating the mAP at IoU thresholds ranging from 0.5 to 0.95 with a step size of 0.05, and then providing the final average value.

*B. Dataset*

The VisDrone2019 dataset is a significant collection of aerial images captured by drones, developed collaboratively by the Machine Learning and Data Mining Laboratory at Tianjin University and the AISKYEYE Data Mining Team. This dataset comprises 288 video clips totaling 261,908 frames and 10,209 static images. These images were taken by cameras mounted on various drones, showcasing diverse scenes across more than a dozen cities in China. The dataset is exceptionally rich, covering a broad range of geographical locations, environmental backgrounds, and object types. Geographically, it includes imagery from 14 different cities in China, providing comprehensive coverage from urban to rural landscapes.

It encompasses multiple types of objects such as pedestrians, cars, bicycles, and more. Additionally, the dataset covers areas with varying population densities, ranging from sparse to densely crowded regions, and was captured under various lighting conditions, including both daytime and nighttime scenes. A notable characteristic of the VisDrone2019 dataset is its inclusion of a large number of small objects of varying sizes depicted at different angles across diverse scenes. This diversity makes the dataset more complex and challenging compared to other computer vision datasets.

*C. Ablation study*

To facilitate the recording of comparative experiments, we have assigned the following naming conventions to the models with various improvements: YOLOv8s serves as the base model and is labeled as Model ID 1.The model that directly adds a small object detection head is labeled as Model ID 2.Model ID 3 is derived from Model 2 by removing the large detection layer to balance performance. Model ID 4 is based on Model 3 and incorporates the Faster-C2f structure. Model ID 5 is an improvement of Model 4 with an enhanced upsampling method.Model ID 6 is derived from Model 5 by adding the EMA attention mechanism.

In this study, YOLOv8s was selected as the baseline model for investigation and further enhancements. The model was trained on the VisDrone dataset using an NVIDIA RTX 4090 GPU (24 GB) on Linux, utilizing PyTorch 1.13 and CUDA 11.6. The experiments primarily rely on the Ultralytics library, version 8.3.18, with a Python environment of 3.9.13. Training involved optimizing key parameters, running for 200 epochs with the Stochastic Gradient Descent (SGD) optimizer set to a momentum of 0.937. The initial learning rate started at 0.01. The learning rate is dynamically adjusted using warm-up and cosine annealing strategies. A batch size of 16 was chosen for efficient memory usage and stable training, with input images resized to 640x640 pixels. A weight decay of 0.0005 was also applied to prevent overfitting and improve model generalization.

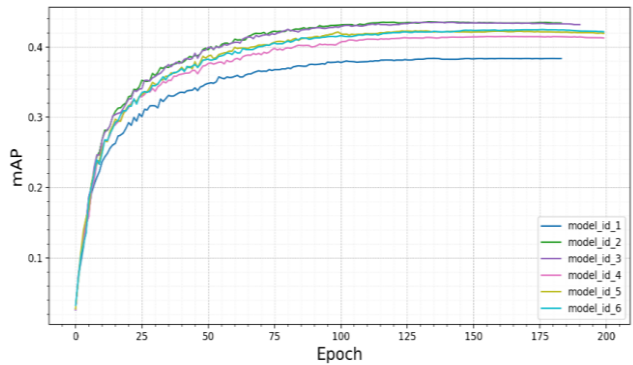

Figure 6 Model training with mAP trend.

To ensure a fair comparison of the impact of model structures on the algorithm, all training strategies remain consistent. Figure 6 illustrates the changes in the model's mAP as the training epochs progress. From the graph, it can be observed that all models have reached a state of convergence.

TABLE 1: Accuracy Performance of Different Innovations on the VisDrone Validation Set

| Model ID | P | R | mAP0.5 |
|---|---|---|---|
| 1(YOLOv8s) | 0.490 | 0.376 | 0.384 |
| 2(Add small target layer) | 0.539 | 0.413 | 0.436 |
| 3(Remove big target layer) | 0.530 | 0.424 | 0.436 |
| 4(Add fast-c2f) | 0.528 | 0.405 | 0.416 |
| 5(Dynamic upsample) | 0.528 | 0.408 | 0.423 |
| 6(Add EMA) | 0.519 | 0.413 | 0.425 |

Model 1 serves as the baseline method, where the model achieves the lowest accuracy. By introducing a small object detection layer, the overall accuracy improves to 43.6%. Model 3 demonstrates that removing the large object detection layer has no impact on detection performance, allowing for a speed improvement without any sacrifice. Model 4 builds on Model 3 by incorporating PConv, which results in a slight decline in mAP performance, a trade-off made to achieve a more lightweight model. The strategies added in Model 5 and Model 6 prove beneficial, as they contribute to an improvement in mAP0.5 on the validation set.

We evaluate the accuracy of the proposed improvements using Precision, Recall, and mAP0.5. Additionally, we assess the model's deployment advantages based on parameter count and computational complexity. TABLE 1 presents the results related to model accuracy, while TABLE 2 showcases the deployment-related outcomes. All comparisons are derived from the proposed enhancements.

TABLE 2: Different Model Innovations with Parameters and GFLOPS

| Model ID | Parameters | GFLOPS |
|---|---|---|
| 1(YOLOv8s) | 11129454 | 28.5 |
| 2(Add small target layer) | 10629048 | 36.7 |
| 3(Remove big target layer) | 7402734 | 34.1 |
| 4(Add fast-c2f) | 6841070 | 30.7 |
| 5(Dynamic upsample) | 6869838 | 30.7 |
| 6(Add EMA) | 6870734 | 31.0 |

Taking both TABLE 1 and TABLE 2 into consideration, it can be concluded that FDM-YOLO achieves a favorable balance between deployment convenience and inference performance. Compared to the baseline model, FDM-YOLO incurs only a minimal increase in computational load while reducing the parameter count by 40%. Additionally, it delivers a 4 percentage point improvement in mAP@0.5.

We also compared FDM-YOLO with other commonly SOTA models. All experiments were conducted using the same training strategy, with the input image size set to 640 pixels. The YOLO series includes models of various sizes, and all tests were conducted using a model size similar to YOLOv8s.

TABLE 3: Accuracy Performance of Other SOTA Model on the VisDrone Validation Set

| Model Name | P | R | mAP0.5 |
|---|---|---|---|
| YOLOv5 | 0.488 | 0.373 | 0.380 |
| YOLOv6 | 0.479 | 0.356 | 0.364 |
| YOLOv8 | 0.490 | 0.376 | 0.384 |
| YOLOv9 | 0.499 | 0.388 | 0.393 |
| YOLOv10 | 0.491 | 0.370 | 0.381 |
| YOLOv11 | 0.507 | 0.377 | 0.386 |
| RT-DETR | 0.432 | 0.247 | 0.221 |
| **FDM-YOLO** | **0.519** | **0.413** | **0.425** |

TABLE 3 demonstrates that, in terms of inference accuracy, our proposed method achieves the best overall performance in small object detection.

The parameters related to Table 4 primarily focus on the lightweight deployment of the model. As shown in the table, our model has the smallest parameter count. Additionally, in terms of inference speed, FDM-YOLO demonstrates strong competitiveness. Overall, FDM achieves the advantages of high inference accuracy, fast inference speed, and a low parameter count, making it a powerful model for small object detection in low-computing scenarios.

TABLE 4: Comparison of Inference Performance Between FDM-YOLO and Other Models

| Model Name | Paramter | Time/ms |
|---|---|---|
| YOLOv5 | 9115406 | 6.9 |
| YOLOv6 | 16299374 | 6.3 |
| YOLOv8 | 11129454 | **5.0** |
| YOLOv9 | 7170958 | 9.8 |
| YOLOv10 | 8042700 | 9.0 |
| YOLOv11 | 9416670 | **5.0** |
| RT-DETR | 32004290 | 12.2 |
| FDM-YOLO | **6870734** | 6.3 |

*D. Visualization*

In this section, we present the visualization results of FDM-YOLOv8 for small object detection in low-computing scenarios across multiple settings, highlighting its advantages over the baseline YOLOv8 model. All images used in this section are sourced from the test set of the VisDrone dataset.

In scenarios with fewer detection objects, specifically in the detection results of small targets that are simple and common, we have concatenated and compared the original image, the image after YOLOV8s detection, and the image after FDM-YOLO detection in sequence, as illustrated in the figure.

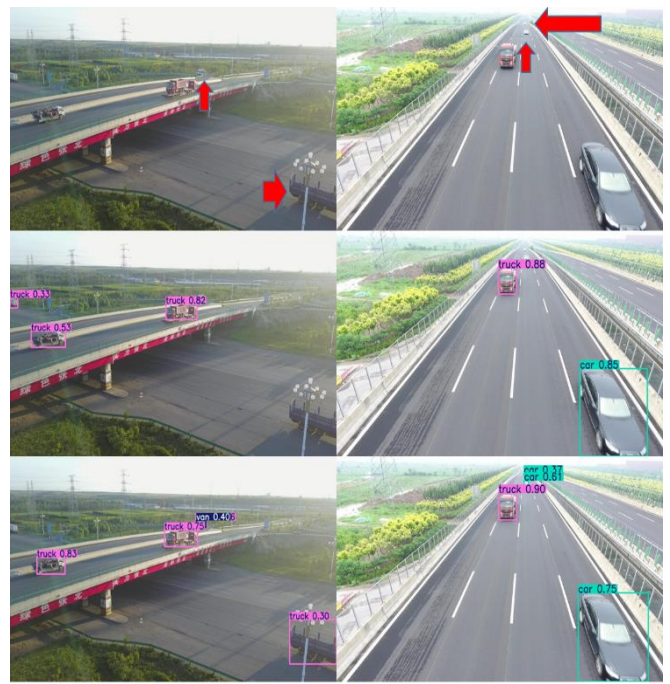

Figure 7 Comparison between YOLOV8s (middle) and FDM-YOLO (bottom) in a simple scenario.

On the left side of the overall image, the YOLOV8s model failed to detect the truck at the bottom right corner, whereas FDM-YOLO successfully identified it. In the middle of the image, there is a very small target that YOLOV8s missed, but FDM-YOLO managed to detect it.

As for the simpler image on the right, there are two tiny car targets located towards the upper middle. These two targets eluded detection by YOLOV8s; however, the FDM-

YOLO model was able to capture both the category and positional information of these small targets.

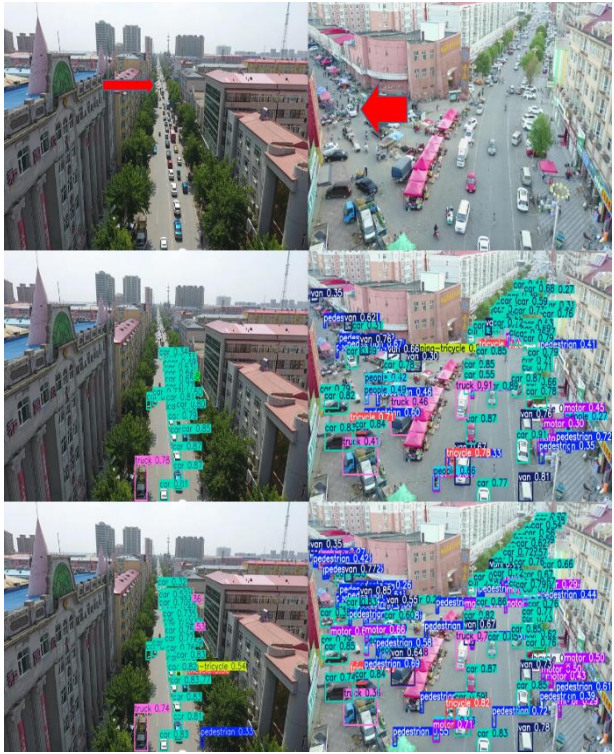

Figure 8 Comparison between YOLOV8s (middle) and FDM-YOLO (bottom) in a dense detection scenario.

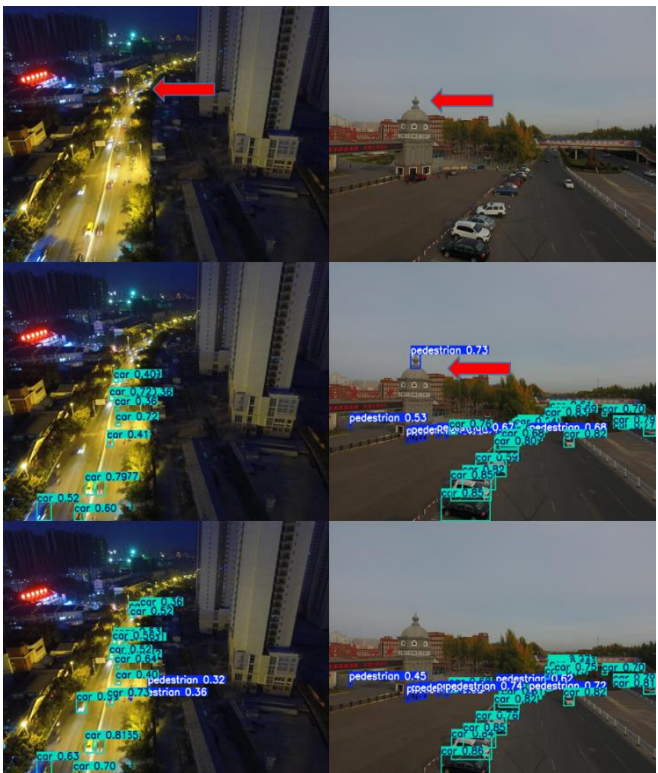

Figure 9 Comparison of detection between YOLOV8 (middle) and FDM-YOLO (bottom) in low-light conditions.

Figure 8 illustrates the comparison of detection performance between YOLOV8 and FDM-YOLO in a dense detection scenario. In the left image, it can be observed that the YOLOV8 model fails to detect small targets at the farthest end of the image. FDM-YOLO effectively mitigates this issue, although the farthest targets remain undetected. At the lower part of the image, FDM-YOLO successfully detects a pedestrian that YOLOV8 misses. As for the right image, the objects to be detected are highly concentrated. The detection density of the FDM-YOLO model is notably higher than that of YOLOV8.

For the image on the left, it is intuitively evident that the detection density of FDM-YOLO (bottom) surpasses that of YOLOV8 (middle) under nighttime conditions. In the central portion of the image, there are some pedestrians that neither model successfully detected, but the overall detection rate for pedestrians is better with the FDM-YOLO model.

Regarding the image on the right, it can be observed that both YOLOV8 and FDM-YOLO exhibit strong competitiveness under low illumination. However, YOLOV8 produced a false detection in the central building area of the image, whereas FDM-YOLO did not manifest such an obvious error.

V. CONCLUSION

Detecting small-scale objects in traffic scenarios presents significant challenges that can reduce overall detection effectiveness. To address these issues, we propose FDM-YOLO, a specialized object detection model designed for aerial photography and traffic scenes dominated by small objects. Built upon YOLOv8, this model focuses on small object detection, enhances feature fusion capabilities, and improves precise localization performance, all without significantly increasing additional computational overhead.

The FDM-YOLO model outperforms widely used models such as YOLOv6 and YOLOv7 across various evaluation metrics. Compared to YOLOv8s, our efficient model significantly enhances object detection performance without substantially increasing computational cost or detection time. It improves recall from 37.6% to 41.3%, precision from 49.0% to 51.9%, and mAP0.5 from 38.0% to 42.5%. Even under challenging conditions such as poor lighting or crowded backgrounds, FDM-YOLO achieves higher IoU values and detects more small objects than YOLOv8s. These capabilities make it highly suitable for applications in UAV-based traffic monitoring.